\documentclass[runningheads]{llncs}
\usepackage{graphicx}
\usepackage{subcaption}
\usepackage{amsmath}
\usepackage{algorithm}
\usepackage{algpseudocode}
\usepackage{multirow}
\usepackage{array}
\newcolumntype{K}{>{\centering\arraybackslash}m{0.12\textwidth}}
\newcolumntype{J}{>{\centering\arraybackslash}m{0.10\textwidth}}
\newcolumntype{L}{>{\centering\arraybackslash}m{0.08\textwidth}}
\newcolumntype{Z}{>{\centering\arraybackslash}m{0.06\textwidth}}

\usepackage{wrapfig}
\usepackage[misc,geometry]{ifsym}

\begin{document}
\title{DeepCPCFG: Deep Learning and Context Free Grammars for End-to-End Information Extraction}
\titlerunning{DeepCPCFG for End-to-End Information Extraction}
%
\author{Freddy C. Chua(\Letter)\orcidID{0000-0002-9809-1152} \and Nigel P. Duffy\orcidID{0000-0003-2663-1647}}
\authorrunning{F. Chua and N. Duffy}
%
\institute{
    Ernst \& Young (EY) AI Lab, Palo Alto, CA, USA \email{freddy.chua@ey.com,nigel.p.duffy@ey.com} 
}

\maketitle              
\begin{abstract}
We address the challenge of extracting structured information from business documents without detailed annotations. We propose Deep Conditional Probabilistic Context Free Grammars (DeepCPCFG) to parse two-dimensional complex documents and use Recursive Neural Networks to create an end-to-end system for finding the most probable parse that represents the structured information to be extracted. This system is trained end-to-end with scanned documents as input and only relational-records as labels. The relational-records are extracted from existing databases avoiding the cost of annotating documents by hand. We apply this approach to extract information from scanned invoices achieving state-of-the-art results despite using no hand-annotations.

\end{abstract}

\section{Introduction}
\label{sec:intro}

Extracting information from business documents remains a significant burden for all large enterprises. Existing extraction pipelines are complex, fragile, highly engineered, and yield poor accuracy. These systems are typically trained using annotated document images whereby human annotators have labeled the tokens or words that need to be extracted. The need for human annotation is the biggest barrier to the broad application of machine learning. 

Business documents often have complex recursive structures whose understanding is critical to accurate information extraction, e.g. invoices contain a variable number of line-items each of which may have multiple (sometimes varying) fields to be extracted. This remains a largely unsolved problem.

Deep Learning has revolutionized sequence modeling resulting in systems that are trained end-to-end and have improved accuracy while reducing system complexity for a variety of applications \cite{NIPS2017_3f5ee243}. Here we leverage these benefits in an end-to-end system for information extraction from complex business documents. This system does not use human annotations making it easily adaptable to new document types while respecting the recursive structure of many business documents.

\begin{figure}[tb]
	\centering
	\scriptsize
	\begin{subfigure}{0.4\columnwidth}
        \centering
		\begin{tabular}{| l r r r |}
			\hline
			& & & \\
			\multicolumn{4}{|r|}{ \fbox{Invoice} \fbox{No:} \fbox{12345} } \\
			\multicolumn{4}{|r|}{ \fbox{Date:} \fbox{12/13/2019} } \\
			& & & \\
			\multicolumn{1}{|c}{ \fbox{Description} }
			&
			\multicolumn{1}{c}{ \fbox{Qty} }
			&
			\multicolumn{1}{c}{ \fbox{Rate} }
			&
			\multicolumn{1}{c|}{ \fbox{Amt} }
			\\
			& & & \\
			\fbox{Apples} & \fbox{5} & \fbox{\$1} & \fbox{\$5} \\
			\fbox{Chicken} \fbox{Wings} & \fbox{10} & \fbox{\$2} & \fbox{\$20} \\
			& & &\\
			\multicolumn{4}{|r|}{ \fbox{Total:} \fbox{\$25} } \\
			& & & \\
			\hline
		\end{tabular}
		\caption{Example of an invoice}
		\label{tbl:invoice_bb}
	\end{subfigure}
	\begin{subfigure}{0.56\columnwidth}
        \centering
		\begin{tabular}{|c c c c c|}
			\hline
			\multicolumn{5}{|l|}{Invoice Relational Table} \\
			\hline
			\textit{InvoiceID} & \multicolumn{2}{c}{\textit{Date}} & \multicolumn{2}{c|}{\textit{TotalAmt}} \\
			12345 & \multicolumn{2}{c}{12/13/2019} & \multicolumn{2}{c|}{\$25} \\
			\hline
			\multicolumn{5}{|l|}{Line-items Relational Table} \\
			\hline
			\textit{InvoiceID} & \textit{Desc} & \textit{Qty} & \textit{Rate} & \textit{Amt} \\
			12345 & Apples & 5 & \$1 & \$5\\
			12345 & Chicken Wings & 10 & \$2 & \$20 \\
			\hline
		\end{tabular}
		\caption{Invoice record in relational tables}
		\label{tbl:record}
	\end{subfigure}
	\caption{Invoice image and invoice record}
\end{figure}

We apply this system to invoices which embody many of the challenges involved: the information they contain is highly structured, they are highly varied, and information is encoded in their 2D layout and not just in their language. A simplified example of an invoice is shown in Figure \ref{tbl:invoice_bb} with the corresponding relational-record shown in Figure \ref{tbl:record}. The relational-record contains structured information including header fields that each appear once in the invoice and an arbitrary number of line-items each containing a description, quantity, rate and amount \footnote{Other business documents are significantly more complicated often involving recursively defined structure.}. While human annotations of document images are rarely available, these relational-records are commonly stored in business systems such as Enterprise Resource Planning (ERP) systems.

Figure \ref{tbl:invoice_bb} illustrates bounding boxes surrounding each of the tokens in the invoice. Some of these bounding boxes, e.g. ``Description'' provide context by which other bounding boxes, e.g. ``Apples'' can be interpreted as corresponding to a description field. We assume that these bounding boxes are provided by an external Optical Character Recognition (OCR) process and given as input to our information extraction pipeline. Unlike alternative approaches which require annotated bounding boxes to train a bounding box classification model, our approach avoids the use of annotations for each box while only requiring relational-records.

A common approach to information extraction is to classify these bounding boxes individually using machine learning and then to use heuristics to group extracted fields to recover the recursive structure. This approach has several shortcomings: 1) It requires annotated bounding boxes, where each annotation is tagged with the coordinates of the box and a known label to train the classification model. 2) It requires heuristic post-processing to reconstruct the structured records and to resolve ambiguities. These post-processing pipelines tend to be brittle, hard to maintain, and highly tuned to specific document types.

Adapting these systems to new document types is an expensive process requiring new annotations and re-engineering of the post-processing pipeline. We address these issues using a structured prediction approach. We model the information to be extracted using a context free grammar (CFG) which allows us to capture even complex recursive document structures. The CFG is built by adapting the record schema for the information to be extracted. Information is extracted by parsing documents with this CFG in 2D. It is important to note that the CFG is not based on the layout of the document, as with previous work in this field, but on the structure of the information to be extracted. We do not have to fully describe the document, we only describe the information that needs to be extracted. This allows us to be robust to a wide variety of document layouts. 

There may be many valid parses of a document. We resolve this ambiguity by extending Conditional Probabilistic Context Free Grammars (CPCFG) \cite{sutton2004conditional} in 3 directions. 1) We use deep neural networks to provide the conditional probabilities for each production rule. 2) We extend CPCFGs to 2D parsing. 3) We train them using structured prediction \cite{10.3115/1118693.1118694}. This results in a computationally tractable algorithm for handling even complex documents.

Our core contributions are: 1) A method for end-to-end training in complex information extraction problems requiring no post-processing and no annotated images or any annotations associated with bounding boxes. 2) A tractable extension of CFGs to parse 2D images where the grammar reflects the structure of the extracted information rather than the layout of the document. 3) We demonstrate state-of-the-art performance in an invoice reading system, especially the problem of extracting line-item fields and line-item grouping.

\section{Related work}
\label{sec:related}

We identify 3 areas of prior work related to ours: information extraction from invoices using machine learning, grammars for parsing document layout, and structured prediction using Deep Learning. The key innovation of our approach is that we use a grammar based on the information to be extracted rather than the document layout resulting in a system that requires no hand annotations and can be trained end-to-end. 

There has been a lot of work on extracting information from invoices. In particular a number of papers investigate using Deep Learning to classify bounding boxes.
\cite{8892875} generated invoice data and used a graph convolutional neural network to predict the class of each bounding box. \cite{layoutlm} uses BERT\cite{devlin-etal-2019-bert} that integrates 2D positional information to produce contextualized embeddings to classify bounding boxes and extract information from receipts\footnote{Receipts are a simplified form of invoice.}.  \cite{majumder-etal-2020-representation} also used BERT and neighborhood encodings to classify bounding boxes.
\cite{denk2019bertgrid,katti-etal-2018-chargrid} use grid-like structures to encode information about the positions of bounding boxes and then to classify those bounding boxes.
\cite{8892875,majumder-etal-2020-representation,layoutlm,yu2020pick} predict the class of bounding boxes and then post-process to group them into records. 
\cite{yu2020pick} uses graph convolutional networks together with a graph learning module to provide input to a Bidirectional-LSTM and CRF decoder which jointly labels bounding boxes. 
This works well for flat (non-recursive) representations such as the SROIE dataset \cite{8977955} but we are not aware of their application to hierarchical structures in more complex documents, especially on recursive structures such as line-items.

\cite{hwang2019post,hwang2020spatial} provide a notable line of work which addresses the line-item problem. \cite{hwang2019post} reduce the 2D layout of the invoice to a sequence labeling task, then use Beginning-Inside-Outside (BIO) Tags to indicate the boundaries of line-items. \cite{hwang2020spatial} treats the line-item grouping problem as that of link prediction between bounding boxes. \cite{hwang2020spatial} jointly infers the class of each box, and the presence of a link between the pair of boxes in a line-item group. However, as with all of the above cited work in information extraction from invoices both \cite{hwang2019post,hwang2020spatial} require hand annotation for each bounding box, and they require post-processing in order to group and order bounding boxes into records. 

Documents are often laid out hierarchically and this recursive structure has been addressed using parsing. 2D parsing has been applied to images \cite{10.1145/2543581.2543593,10.5555/2986459.2986468,10.5555/2981780.2982029}, where the image is represented as a 2D matrix. The regularity of the 2D matrix allows parsing to be extended directly from 1D. The 2D approaches to parsing text documents most related to ours are from \cite{10.1109/ICDAR.2005.98,10.1109/ICCV.2005.140,Tomita1991}. \cite{Tomita1991} described a 2D-CFG for parsing document layout which parsed 2D regions aligned to a grid. \cite{10.1109/ICDAR.2005.98} describe 2D-parsing of document layout based on context free grammars. Their Rectangle Hull Region approach is similar to our 2D-parsing algorithm but yields a $\mathcal{O}(n^5)$ complexity compared to our $\mathcal{O}(n^3)$. \cite{10.1109/ICCV.2005.140} extends \cite{10.1109/ICDAR.2005.98} to use conditional probabilistic context free grammars based on Boosting and the Perceptron algorithm while ours is based on deep Learning with back-propagation through the parsing algorithm. Their work relies on hand-annotated documents.
All of this work requires grammars describing the document layout and seeks to fully describe that layout.  On the other hand, our approach provides end-to-end extraction from a variety of layouts simply by defining a grammar for the information to be extracted and without a full description for the document layout. To the best of our knowledge, no other work in the space of Document Intelligence takes this approach.

\cite{LARI199035} provides the classical Probabilistic Context Free Grammar (PCFG) trained with the Inside-Outside algorithm, which is an instance of the Expectation-Maximization algorithm. This work has been extended in a number of directions that are related to our work. \cite{sutton2004conditional} extended PCFGs to the Conditional Random Field \cite{10.5555/645530.655813} setting. \cite{drozdov-etal-2019-unsupervised-latent} use inside-outside applied to constituency parsing where deep neural networks are used to model the conditional probabilities. Like us they train using backpropagation, however, they use the inside-outside algorithm while we use structured prediction. Other work \cite{AlvarezMelis2017TreestructuredDW} considers more general applications of Deep Learning over tree structures. While \cite{drozdov-etal-2019-unsupervised-latent,LARI199035,sutton2004conditional} are over 1D sequences, here we use deep neural networks to model the conditional probabilities in a CPCFG over 2D structures. 

Finally, we refer the reader to \cite{subramani2020survey} for a recent survey on Deep Learning for document understanding. 

\section{Problem description}
\label{sec:problem}

We define an information extraction problem by a universe of documents $D$ and a schema describing the structure of records to be extracted from each document $d \in D$. Each document is a single image corresponding to a page in a business document (extensions to multi-page documents are trivial). We assume that all documents $d$ are processed (e.g. by OCR software) to produce a set of bounding boxes $b = (content, x_1, y_1, x_2, y_2)$ with top left coordinates $(x_1, y_1)$ and bottom right coordinates $(x_2, y_2)$. 

The schema describes a tuple of named fields each of which contains a value. Values correspond to base types (e.g., an Integer), a list of values, or a recursively defined tuple of named fields. These schemas will typically be described by a JSON Schema, an XML Schema or via the schema of a document database. More generally, the schema is a context free grammar $G=(V, \Sigma, R, S)$ where $\Sigma$ are the terminals in the grammar and correspond to the base types or tokens, $V$ are the non-terminals and correspond to field names, $R$ are the production rules and describe how fields are constructed either from base types or recursively, and $S$ is the start symbol corresponding to a well-formed extracted record. 

\begin{figure}[htb]
	\centering
	\begin{tabular}{|r l l|}
		\hline
		\textbf{Invoice} & \textbf{:=} & \textbf{(InvoiceID Date LineItems TotalAmt)} ! \\
		\textbf{InvoiceID} & \textbf{:=} & \textbf{STRING} $|$ (N InvoiceID) ! $|~\epsilon$ \\
		\textbf{Date} & \textbf{:=} & \textbf{STRING $|$ Date Date} $|$ (N Date) ! $|~\epsilon$ \\
		\textbf{TotalAmt} & \textbf{:=} & \textbf{MONEY} $|$ (N TotalAmt) ! $|~\epsilon$ \\
		\textbf{LineItems} & \textbf{:=} & \textbf{LineItems LineItem} $|$ \textbf{LineItem} \\
		\textbf{LineItem} & \textbf{:=} & \textbf{(Desc Qty Rate Amt)} ! $|$ (N LineItem) ! \\
		\textbf{Desc} & \textbf{:=} & \textbf{STRING $|$ Desc Desc} $|$ (N Desc) !\\
		\textbf{Qty} & \textbf{:=} & \textbf{NUMBER} $|$ (N Qty) ! $|~\epsilon$  \\
		\textbf{Rate} & \textbf{:=} & \textbf{NUMBER} $|$ (N Rate) ! $|~\epsilon$ \\
		\textbf{Amt} & \textbf{:=} & \textbf{MONEY}  $|$ (N Amt) ! $|~\epsilon$ \\
		N & := & N N $|$ STRING \\
		\hline
	\end{tabular}
	\caption{Grammar}
	\label{tbl:grammar}
\end{figure}

An example grammar is illustrated in Figure \ref{tbl:grammar}. Reading only the content in \textbf{bold} gives the rules for $G$ that represents the main information that we want to extract. Here $\Sigma=\{$STRING, NUMBER. MONEY$\}$, $V=\{$Invoice, InvoiceNumber, TotalAmount, LineItems, LineItem, Desc (Description), Qty (Quantity), Rate, Amt$\}$ \footnote{Note that this is not in CNF but the conversion is straightforward \cite{cole2007converting}.}. The goal of information extraction is to find the parse tree $t_d \in G$ corresponding to the record of interest.

\section{Approach}
\label{sec:approach}
We augment $G$ to produce $G'=(V', \Sigma', R', S')$ a CPCFG whose parse trees $t'_d$ can be traversed to produce a $t_d \in G$. Below we assume (without loss of generality) that all grammars are in Chomsky Normal Form (CNF) and hence all parse trees are binary. 

The set of bounding boxes for a document $d$ may be recursively partitioned to produce a binary tree. We only consider partitions that correspond to vertical or horizontal partitions of the document region so that each subtree corresponds to a rectangular region $B$ of the document image \footnote{This approach is built on the assumption (by analogy with sentences) that documents are read contiguously. This assumption likely does not hold for general images where occlusion may require non-contiguous scanning to produce an interpretation.}. Each such region contains a set of bounding boxes. We consider any two regions $B_1$ and $B_2$ equivalent if they contain exactly the same set of bounding boxes.

The leaves of a partition tree each contain single bounding boxes. We extend the tree by appending a final node to each leaf that is the bounding box itself. We refer to this extension as the partition tree for the remainder of the paper. The contents of these bounding boxes are mapped to the terminals 
$\Sigma'=\Sigma \cup \{$NOISE$, \epsilon\}$. The special NOISE token is used for any bounding box contents that do not map to a token in $\Sigma$. The special $\epsilon$ token is used to indicate that some of the Left-Hand-Side (LHS) non-terminals can be an empty string. Document fields indicated with $\epsilon$ are optional and can be missing from the document. We handle this special case within the parsing algorithm.

We augment the non-terminals $V'=V \cup \{$N$\}$ where non-terminal N captures regions of the image that contain no extractable information.

We augment the rules $R$ by adding rules dealing with the N and NOISE symbols. Every rule $X\rightarrow YZ$ is augmented with a production $X \rightarrow NX$ and every rule $A \rightarrow \alpha$ is augmented with a rule $A \rightarrow NA$. In many cases the record information in a document may appear in arbitrary order. For example, the order of line-items in an invoice is irrelevant to the meaning of the data. We introduce the suffix ``!'' on a production to indicate that all permutations of the preceding list of non-terminals are valid. This is illustrated in Figure \ref{tbl:grammar} where the modifications are in not in \textbf{bold}.

Leaves of a partition tree are labeled with the terminals mapped to their bounding boxes. We label the internal nodes of a partition tree with non-terminals in $V'$ bottom up. Node $i$ corresponding to region $B_i$ is labeled with any $X \in V'$ where $\exists (X \rightarrow YZ) \in R'$ such that the children of node $i$ are labeled with $Y$ and $Z$. We restrict our attention to document partition trees for which such a label exists and refer to the set of all labels of such trees as $T'_d$ and a single tree as $t'_d$ (with a minor abuse of notation). We recover a tree $t_d \in G$ from $t'_d$ by removing all nodes labeled with N or NOISE and re-connecting in the obvious way. We say that such a $t_d$ is compatible with $t'_d$.

By adding weights to each production, we convert $G'$ to a CPCFG. Trees are assigned a score $s(t'_d)$ by summing the weights of the productions at each node. Here the weights are modeled by a deep neural network $m_{X\rightarrow YZ}(B_1, B_2)$ applied to a node labeled by $X$ with children labeled by $Y$ and $Z$ and contained in regions $B_1$ and $B_2$ respectively.

\subsection{Parsing}
\label{sec:parsing}

\begin{figure}[tb]
    \centering    
    \includegraphics[width=\textwidth]{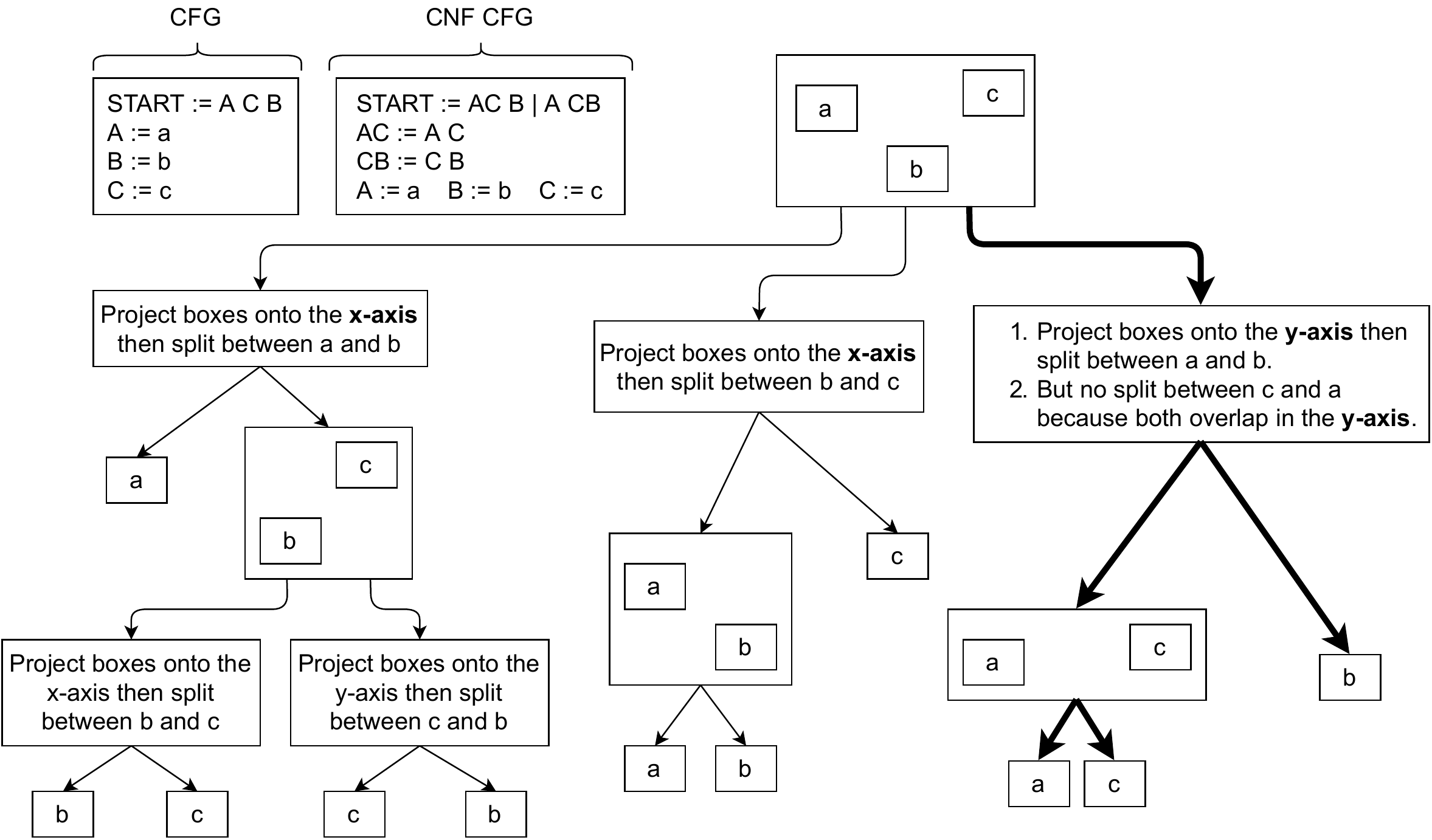}
    \caption{An example of CYK 2D parsing on a simple document and grammar. The valid parse tree is the one on the left with \textbf{bolded arrows} $\rightarrow$.}
    \label{fig:cyk2d}
\end{figure}

We can now solve the information extraction problem by finding the tree $t'_d$ with the highest score by running a chart parser. 

A typical 1D chart parser creates a memo $c[$sentence span$][X]$ where $X \in V$ and ``sentence span'' corresponds to a contiguous span of the sentence being parsed and is usually represented by indexes $i,j$ for the start and end of the span so that the memo is more typically described as $c[i][j][X]$. The memo contains the score of the highest scoring subtree that could produce the sentence span and be generated from the non-terminal $X$. The memo can be constructed (for a sentence of length $n$) top down starting from $c[0][n][S]$ with the recursion:
\begin{align}
    \label{eqn:dp_1}
    c[i][j][X] = \max_{(X \rightarrow YZ) \in R} ~ \max_{i \leq k < j} \Big( w_{X \rightarrow YZ}(i,k,j) + c[i][k][Y] + c[k][j][Z] \Big)
\end{align}
where $w_{X \rightarrow YZ}$ is the weight associated with the rule $X \rightarrow YZ$. It is easy to see that the worst-case time complexity of this algorithm is $O(n^3 |R'|)$.

We extend this algorithm to deal with 2D documents. In this case the memo $c[B][X]$ contains the score of the highest scoring sub-tree that could produce the region $B$ of the document image from the non-terminal $X$. This results in a top down algorithm recursively defined as follows:
\begin{align}
    \label{eqn:dp_2}
    c[B][X] = \max_{(X \rightarrow YZ) \in R'} ~ \max_{B_1, B_2 \in Part(B)} \Big(& m_{X\rightarrow YZ}(B_1, B_2) \\ \nonumber
    &\quad + c[B_1][Y] + c[B_2][Z] \Big)
\end{align}
where we consider $Part(B)$ defined as partitions of $B$ obtained by splitting horizontally or vertically between adjacent pairs of bounding boxes in $B$, i.e. $B = B_1 \cup B_2, B_1 \cap B_2 = \emptyset$. There are $n-1$ such horizontal splits and $n-1$ such vertical splits. The worst-case time complexity of this algorithm is $O(n^3 |R'|)$ \footnote{The cost of calculating $c$ can be further reduced by using beam search.}.

Overloading notation, we say $s(d) = c[d][S']$ provides the score for the highest scoring tree for $d$. We can recover the tree itself by maintaining back-pointers as in a typical chart parser. We assign a score to $t_d \in G$ as the maximum score over all $t'_d$ with which it is compatible.

One may refer to Figure \ref{fig:cyk2d} for an illustration of how 2D parsing is done using the CYK algorithm. 

\subsection{Learning}
\label{sec:learning}

\begin{algorithm}[tb]
    \caption{The neural network model of the unary rules}
    \label{alg:f_unary}
    \begin{algorithmic}
    \Function{$m_{X \rightarrow x}$}{$bb, lm$}
        \Comment{$bb$: the bounding box, $lm$: the language model}
        \State $h_0$ = forward($lm, bb$) \Comment{Get language model vector}
        \State $h$ = GELU($W_0^{X \rightarrow x} h_0 + b_0^{X \rightarrow x}$) \Comment{Hidden vector}
        \State $s$ = $W_1^{X \rightarrow x} h + b_1^{X \rightarrow x}$ \Comment{Score for Non-terminal $X$}
        \State \Return $s, h$
    \EndFunction
    \end{algorithmic}
\end{algorithm}
\begin{algorithm}[tb]
    \caption{The neural network model of the binary rules}
    \label{alg:f_binary}
    \begin{algorithmic}
        \Function{$m_{X \rightarrow YZ}$}{$B, B_1, B_2, h_d, M$}
	            \Comment{$B$: the region under consideration.}\\
	            \Comment{$B_1, B_2$: the sub-regions for one of the partitions of $B$.}\\
	            \Comment{$h_d$: the embedding representing the direction of the partition.}\\
	            \Comment{$M$: the memoization of the dynamic program.}
	            \State $h_1, h_2 = M[B_1][Y][h], M[B_2][Z][h]$
	            \State $h$ = GELU($W_1^{X \rightarrow YZ} h_1 + W_2^{X \rightarrow YZ} h_2 + W_d^{X \rightarrow YZ} h_d + b_0^{X \rightarrow YZ}$) \Comment{Hidden vector}
	            \State $s = W_3^{X \rightarrow YZ} h + b_3^{X \rightarrow YZ}$ \Comment{Score for Non-terminal $X$}
	            \State \Return $s, h$
        \EndFunction
    \end{algorithmic}
\end{algorithm}

Given training data consisting of pairs $(d, t_d)$ with $d \in D$ a document and $t_d \in G$ our goal is to learn the parameters of the models $m_r$ such that $t_d$ is the highest scoring tree for $d$.

We achieve this following the structured prediction approach  \cite{10.3115/1118693.1118694} and minimizing the structured prediction loss.
\begin{equation}
	\label{eqn:obj2}
	\sum_{d \in D} s(\hat{t'_d}) - s(\bar{t'_d})
\end{equation}
Where $\bar{t'_d}$ is the highest scoring tree compatible\footnote{We will address compatibility in Section \ref{sec:compatibility}.} with $t_d$ (the correct tree) and $\hat{t'_d}$ is the highest scoring tree from the dynamic program. Intuitively we aim to increase the scores of correct trees and decrease scores of incorrect trees.

We perform this minimization using gradient descent and back-propagation on Equation \ref{eqn:obj2}. Each model $m_r$ is a deep neural network and the score $s(t'_d)$ is computed recursively as a function of these models. We can back-propagate through this recursion to jointly train all of the $m_r$.

It remains to describe the models $m_r$. Each model outputs both a score for the production at a given tree node and an embedding meant to represent the sub-tree under that node. The models for terminal rules $X \rightarrow x$ where $x \in \Sigma$ take as input a target bounding box and any context that might be relevant to labeling that bounding box such as the coordinates of the bounding box. Intuitively these models predict which Non-terminal labels a given bounding box. For simplicity of presentation, we describe a relatively simple class of models. 

Algorithm \ref{alg:f_unary} shows the model $m_{X \rightarrow x}$ used in Equation \ref{eqn:dp_1}. The forward function in Algorithm \ref{alg:f_unary} produces a vectorized representation of the given bounding box. One could either use a language model with pretrained weights or train the model end-to-end as part of the learning process. Many architectures are possible for such a model including ones based on language embeddings (e.g., BERT \cite{devlin-etal-2019-bert}) that embed only the contents of the bounding box, and ones which aim to take document image, layout, and format into account (e.g., LayoutLM \cite{layoutlm}). 

When we integrate our DeepCPCFG model and a language model, it results in an encoder-decoder architecture. The encoder is the language model (e.g. Layoutlm) which takes as input the bounding box coordinates and text, then produces a word embedding for each bounding box. The word embedding is given to the decoder, which is DeepCPCFG, that produces a parse tree reflecting the document hierarchy. Geometric information is captured explicitly by Layoutlm then implicitly again during 2D parsing. 

The models for Non-terminal rules $X \rightarrow YZ$ take as inputs the sub-regions $B_1, B_2 \in Part(B)$ such that $B = B_1 \cup B_2, B_1 \cap B_2 = \emptyset$ where $B, B_1, B_2$ are labeled with $X, Y, Z$ respectively and outputs a score for $X$ and an embedding vector for $B$ representing $X$.

Algorithm \ref{alg:f_binary} shows the implementation of the function $m_{X \rightarrow YZ}$ (used in Equation \ref{eqn:dp_2}), as a derivation of a Tree Convolutional Block \cite{harer2019treetransformer}. The matrices $W_i^{r}$ with biases $b_i^r$ are the learned parameters of each model $m_r$ and $M[B][X][h]$ is the embedding for the best scoring sub-tree associated with $B$ and generated by $X$. These embedding vectors are stored in the memo $M$ of the chart parser.

\subsection{Compatible tree for structured prediction}
\label{sec:compatibility}

We showed in Equation \ref{eqn:obj2} of Section \ref{sec:learning} that a compatible tree $\bar{t'_d}$ with the annotation of $d$ is required as part of learning the parameters for a Structured Prediction model. We define a tree $\bar{t'_d}$ to be most \emph{compatible} if the tree gives the \emph{smallest} tree edit-distance \cite{zhang1989simple} as compared with the hierarchical structure derived from the relational-record of the document $d$. But ordering of columns (tree branches) within a relational database can interfere with how tree edit-distance is computed. Therefore, we propose to relax the ordering of fields within the document, and the fields within the recurrent line-items to derive a variant of tree edit-distance, which we call Hierarchical Edit-Distance (HED). In HED, we only require the ordering of line-items within a document and words within a field remain the same, while the ordering of fields within a line-item may be permuted without impacting the distance.

\begin{align}
    \label{eqn:hed}
    \operatorname{HED}(x, y) = \sum_{f \in H} \operatorname{SED}(x_f, y_f) + \operatorname{LiSeqED}(x_{\text{li}}, y_{\text{li}})
\end{align}
$H$ refers to the set of header fields: \{InvoiceID, Date, TotalAmt\}. SED stands for Levenshtein String Edit Distance. LiSeqED (Line-item sequence edit-distance) is defined by Equation \ref{eqn:liseqed}. $x_{\text{li}}$ and $y_{\text{li}}$ represent the line-items of $x$ and $y$.

\begin{align}
    \label{eqn:liseqed}
    \operatorname{LiSeqED}(x, y) = \begin{cases}
        \sum_{i=1}^{|x|} \operatorname{LiED}(x_{i}, \emptyset) \quad \text{if} ~ |y| = 0 \\
        \sum_{i=1}^{|y|} \operatorname{LiED}(\emptyset, y_{i}) \quad \text{if} ~ |x| = 0 \\
        \text{otherwise:} \\
        \min \begin{cases}
            \operatorname{LiED}(x_1, y_1) + \operatorname{LiSeqED}(\operatorname{tail}(x), \operatorname{tail}(y)) \\
            \operatorname{LiED}(x_1, \emptyset) + \operatorname{LiSeqED}(\operatorname{tail}(x), y) \\
            \operatorname{LiED}(\emptyset, y_1) + \operatorname{LiSeqED}(x, \operatorname{tail}(y)) \\
        \end{cases} 
    \end{cases}
\end{align}
where LiED is defined by Equation \ref{eqn:lied}, tail is a function that returns the rest of the list except the first element in the list. $\emptyset$ represents an empty line-item.

\begin{align}
    \label{eqn:lied}
    \operatorname{LiED}(x, y) = \sum_{f \in G} \operatorname{SED}(x_f, y_f)
\end{align}
where $G$ represents the set of line-item fields \{Desc, Qty, Rate, Amt\}.

Using HED, we can obtain the compatible (smallest edit-distance) tree $\bar{t'_d}$ in the same way as we obtained the highest scoring parse tree $\hat{t'_d}$ by using HED as the scoring function then taking the minimum instead of maximum in Equations \ref{eqn:dp_1} and \ref{eqn:dp_2}. We will re-use HED when evaluating the results of our experiments.

\section{Experiments}
\label{sec:experiments}

\begin{figure}[tb]
    \centering
    \begin{subfigure}[b]{0.3\textwidth}
        \includegraphics[width=\textwidth]{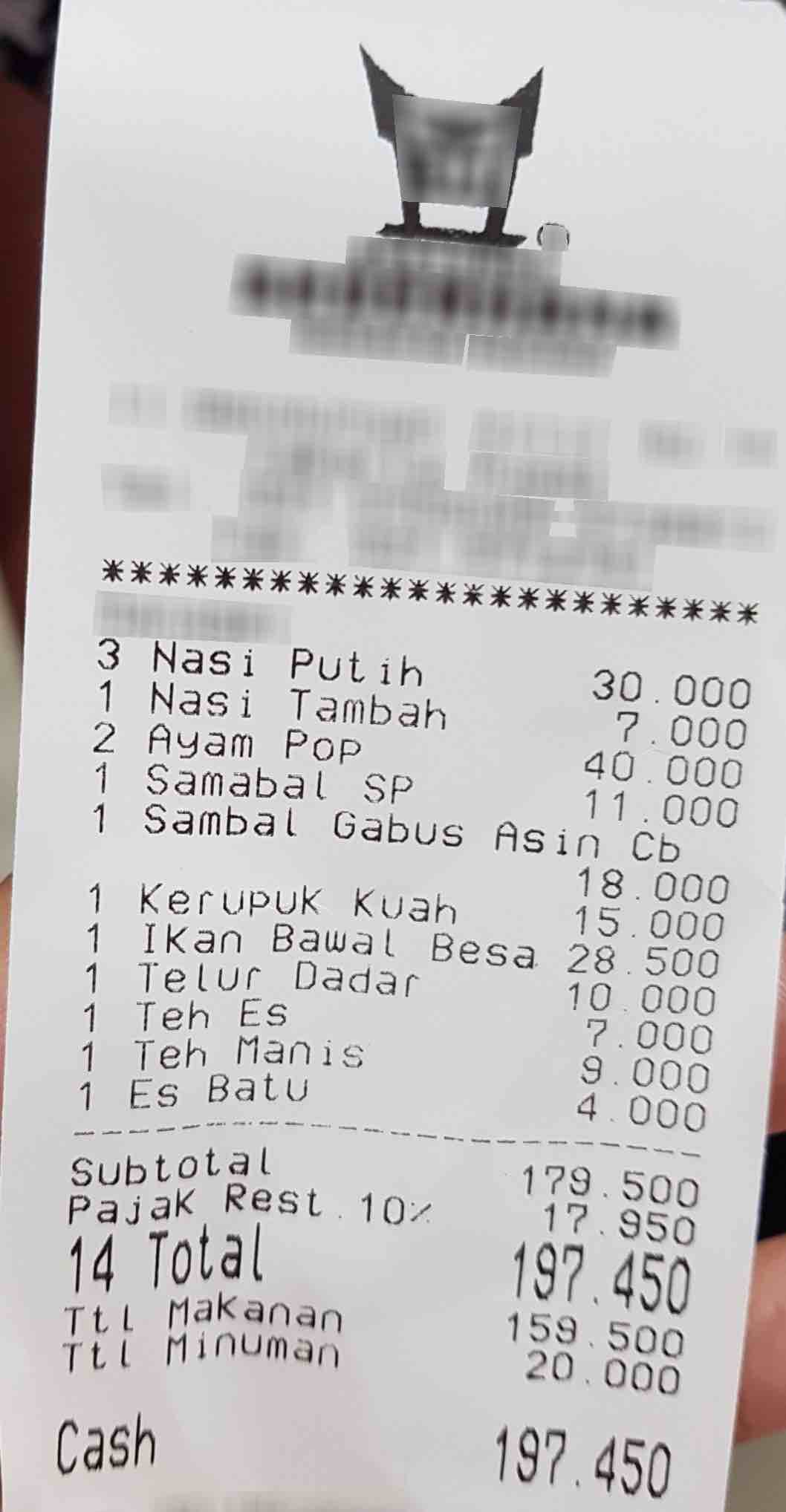}
        \caption{CORD receipt}
        \label{fig:cord-example1}
    \end{subfigure}
    \begin{subfigure}[b]{0.6\textwidth}
        \includegraphics[width=\textwidth]{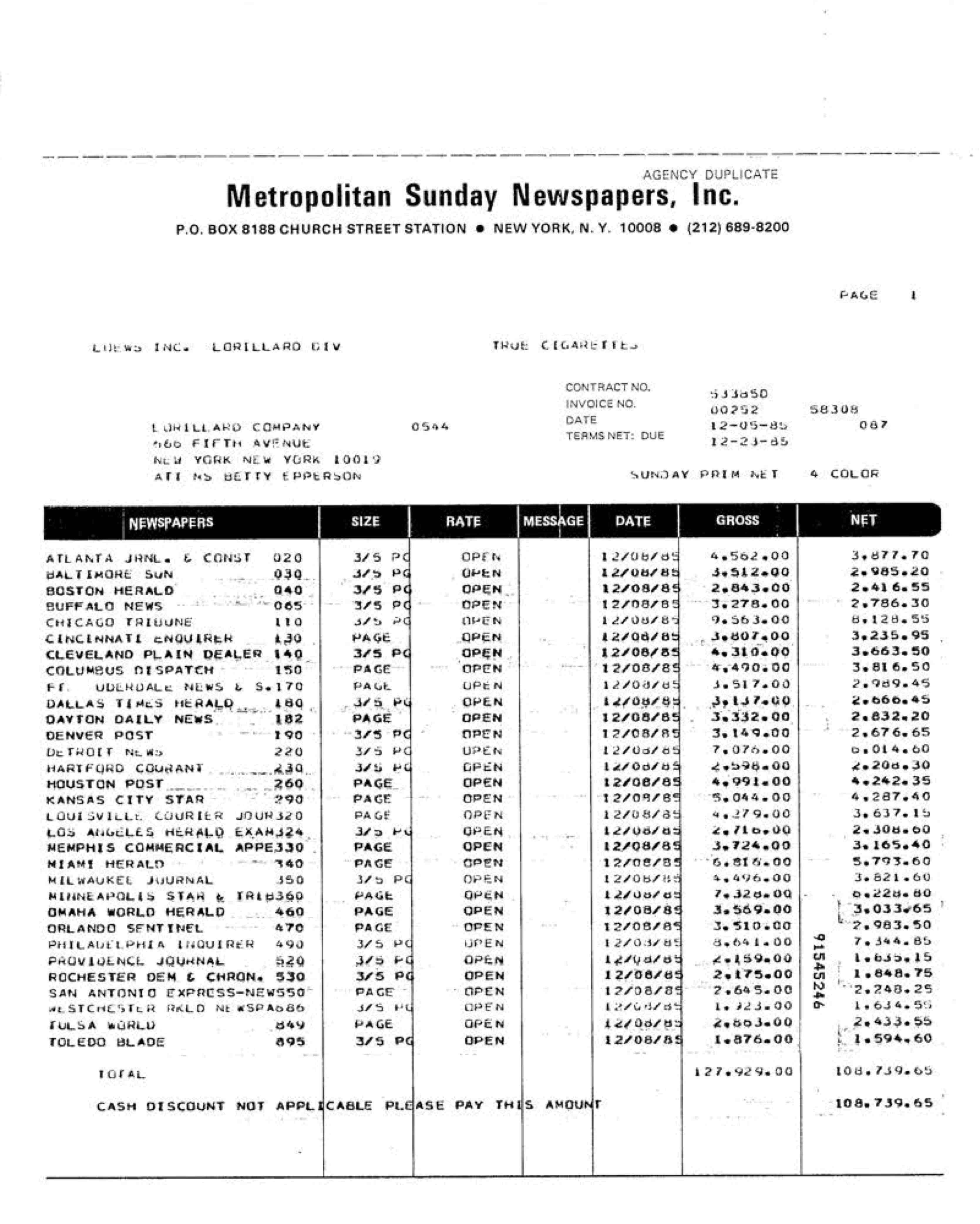}
        \caption{RVL-CDIP invoice}
        \label{fig:ARIA_CDIP_Test_91545246}
    \end{subfigure}
    \caption{Examples of structured documents}
    \label{fig:example_docs}
\end{figure}

Prior research on business documents, particularly invoices, has been limited by available public datasets. We are not aware of public datasets that provide structured extraction from business documents or invoices. In the FUNSD dataset \cite{jaume2019} annotations reflect record linkages rather than hierarchical structures. The RVL-CDIP \cite{harley2015icdar} dataset provides class labels for a document classification task rather than line-item annotations for information extraction. \cite{Gralinski2020KleisterAN} provides a dataset of business documents for information extraction but its documents do not contain line-items. The receipts used in SROIE \cite{8977955} have line-items but those are not annotated. \cite{denk2019bertgrid,katti-etal-2018-chargrid,yu2020pick,majumder-etal-2020-representation} report results only on proprietary sets of invoices.


To evaluate our model on line-items we ran our experiments on three datasets summarized in Table \ref{tbl:dataset}. The first dataset from \cite{hwang2019post,hwang2020spatial} consists of the CORD receipt data. The second consists of proprietary invoices for which we have both hand annotations and relational-records. We created the third dataset by hand annotating invoices in the RVL-CDIP collection \cite{harley2015icdar}. We use the RVL-CDIP invoices solely for testing and welcome other researchers to compare with our results \footnote{We release RVL-CDIP data and metric code at \url{https://github.com/deepcpcfg}}.

\begin{table}[t]
    \centering
    \caption{Dataset sizes}
    \label{tbl:dataset}
    \begin{tabular}{|c|c|c|c|}
        \hline
        \multirow{2}{*}{} & Receipts & \multicolumn{2}{c|}{Invoices} \\
        \cline{2-4}
        & CORD & Proprietary & RVL-CDIP \\
        \hline
        Training & 800 & 17938 & 0 \\
        Validation & 100 & 2085 & 0 \\
        Testing & 100 & 2516 & 869 \\
        \hline
    \end{tabular}
\end{table}

Our preliminary experiments demonstrated that the language model used in Algorithm \ref{alg:f_unary} should be fine-tuned on relevant documents. 
In the experiments below we fine-tune Layoutlm\cite{layoutlm} as follows: 
using the untrained CFG we derive compatible trees from the proprietary invoices' relational-records. For each invoice $d$, we obtain the compatible tree $\bar{t'_d}$, which provides classes for each bounding box (token) from the leaves of the compatible trees. These leaves are used to fine-tune a LayoutLM model on the token classification task. 

\subsection{Results on CORD}
Table \ref{tbl:cord} provides a comparison between \cite{hwang2020spatial} and DeepCPCFG. We compare against the best results reported in Table 9 of \cite{hwang2020spatial} using the SPADE metric described in the appendix of \cite{hwang2020spatial}\footnote{At the time of writing, \cite{hwang2020spatial} have not released their code. We re-implemented SPADE metric to the best of our understanding based on communication with the authors. We release the metric implementation and output files at \url{ https://github.com/deepcpcfg} for anyone to compare or verify our results.}.
The SPADE (Spatial Dependency Parsing) metric can be seen as a special case of HED if SED in Equations \ref{eqn:hed} and \ref{eqn:lied} is implemented using exact string match as follows,
\begin{align}
    SED_{SPADE}(x,y) = 
    \begin{cases}
        0 & \quad \text{if} ~ x = y \\
        1 & \quad \text{if} ~ x \neq y
    \end{cases}
\end{align}

Using the dataset provided by \cite{hwang2020spatial} we derive relational-records (see Figure \ref{tbl:record}) for our DeepCPCFG model. In table \ref{tbl:cord}, \cite{hwang2020spatial} is trained using hand annotations while DeepCPCFG makes no use of those annotations and is therefore at a significant disadvantage. Although DeepCPCFG can leverage hand annotations when they are present, we chose not to use them to emphasize the power of DeepCPCFG when trained end-to-end.

Overall DeepCPCFG achieves comparable results to \cite{hwang2020spatial} despite not being trained on hand annotations. Given that there are only 100 receipts in the holdout test set, the numbers we report depend substantially on 1 or 2 receipts.
When we inspect DeepCPCFG's errors we found rotated or distorted receipts (see Fig \ref{fig:cord-example1}). DeepCPCFG's performance is compelling given that DeepCPCFG focuses on learning an end-to-end model for formally scanned business documents while \cite{hwang2020spatial} specializes in extraction from photos taken using handheld cameras.

\begin{table}[tb]
    \centering
    \caption{F1 results on CORD using SPADE metric}
    \label{tbl:cord}
    \begin{tabular}{|m{0.25\columnwidth}|J|L|L|L|L|K|}
        \hline
        Model & {\bf Overall} & Desc & Qty & Rate & Amt & TotalAmt \\
        \hline
        Hwang et al. \cite{hwang2020spatial} & 90.1 & \textbf{91.6} & \textbf{92.1} & 91.6 & \textbf{93.4} & \textbf{96.9} \\ 
        \hline
        DeepCPCFG & \textbf{92.2} & 88.7 & 90.6 & \textbf{96.4} & 91.7 & 95.0 \\ 
        \hline
    \end{tabular}
\end{table}

\subsection{Results on invoices}
Scanned invoices reflect the primary objective and motivation for our research, that is, information extraction from complex business documents. By comparison, the CORD receipt dataset is much smaller and simpler. 

In Table \ref{tbl:hed-proprietary-rvl-cdip} we report results of 3 models trained on our proprietary invoices. The models are evaluated on a holdout test set of proprietary invoices and on an unseen set of invoices from RVL-CDIP. In all cases we train only on relational-records and do not make use of hand annotations.

For invoices, we report results based on HED which allows for mismatches due to OCR errors.
HED is implemented by tracking the number of unchanged characters (true positives), insertions (false negatives) and deletions (false positives) required to transform each prediction into its respective annotation. Replacements are treated as deletion/insertion pairs. These counts then allow us to derive precision, recall and f1 metrics.

First, we investigate DeepCPCFG with pre-trained Layoutlm as the language model. We compare this to DeepCPCFG using fine-tuned Layoutlm and note a significant improvement in performance on both datasets.\footnote{In each case the optimal number of epochs was chosen using the validation set.}

We also examine the performance of untrained DeepCPCFG (Epoch 0) and observe that training DeepCPCFG provides a dramatic improvement in performance especially for the Desc field which is the most dependent on structure as it is composed of multiple bounding boxes.

\begin{table}[tb]
    \centering
    \caption{F1 results on proprietary invoices using HED metric}
    \label{tbl:hed-proprietary-rvl-cdip}
    \begin{tabular}{|m{0.32\columnwidth}|L|Z|Z|Z|Z|J|Z|J|}
        \hline
        \multicolumn{1}{|c|}{Model} & \scriptsize{\bf Overall} & Desc & Qty & Rate & Amt & \scriptsize{InvoiceID} & Date & \scriptsize{TotalAmt} \\
        \hline \hline
        \multicolumn{9}{|c|}{Proprietary Invoices} \\
        \hline
        DeepCPCFG pre-trained LayoutLM (Epoch 3) & 67.2 & 61.7 & 70.0 & 70.7 & 77.2 & 77.8 & 86.2 & 84.1 \\
        \hline
        DeepCPCFG fine-tuned LayoutLM (Epoch 0) & 73.5 & 68.7 & 81.2 & 83.4 & 81.8 & 92.8 & 86.3 & 80.9 \\
        \hline
        DeepCPCFG fine-tuned LayoutLM (Epoch 1) & 82.2 & 79.1 & 86.8 & 89.5 & 88.6 & 93.0 & 88.2 & 83.9 \\
        \hline
        Compatible Trees & 95.6 & 97.3 & 91.5 & 95.2 & 92.8 & 94.8 & 89.6 & 87.5 \\
    \hline \hline
        \multicolumn{9}{|c|}{RVL-CDIP Invoices} \\
        \hline
        DeepCPCFG pre-trained LayoutLM (Epoch 3)  & 55.2 & 52.0 & 38.4 & 45.6 & 60.0 & 53.6 & 75.2 & 68.1 \\
        \hline
        DeepCPCFG fine-tuned LayoutLM (Epoch 0) & 63.1 & 60.0 & 48.7 & 57.2 & 66.8 & 75.5 & 83.7 & 68.5 \\
        \hline
        DeepCPCFG fine-tuned LayoutLM (Epoch 1) & 70.5 & 69.0 & 55.2 & 57.0 & 73.5 & 74.6 & 84.5 & 71.6 \\
        \hline
        Compatible Trees & 89.9 & 92.3 & 80.0 & 88.2 & 83.3 & 85.2 & 88.5 & 80.7 \\
        \hline
    \end{tabular}
\end{table}

The HED metric is sensitive to OCR or annotation errors. The rows ``Compatible Trees'' in Table \ref{tbl:hed-proprietary-rvl-cdip} show the quality of the compatible trees derived using the relational-records annotations. These values reflect the best results possible on the holdout test set given OCR and annotation errors.

The RVL-CDIP dataset is rather old and its scanned images are typically noisy or of poor quality (see Figure \ref{fig:ARIA_CDIP_Test_91545246}) resulting in diminished OCR quality. This leads to deteriorated performance for the RVL-CDIP dataset in Table \ref{tbl:hed-proprietary-rvl-cdip}.

\subsubsection{Relation to bounding box classification}
While our goal is to extract structured information from complex documents in an end-to-end fashion it is informative to compare against methods that classify bounding boxes based on hand annotations. In Table \ref{tbl:hed-proprietary-rvl-cdip1} we compare DeepCPCFG against a Layoutlm based bounding box classifier. Note that these results measure the number of bounding boxes where the model and the human annotator disagree on their label. 

We first examine the performance of bounding box classifications produced by taking the leaves from the compatible tree (``Leaf-annotations''). This process uses the true relational-records on the test data and identifies the bounding boxes that best recover those records. This performs poorly particularly on fields like Rate, Amt, Date, and TotalAmt whose values may appear multiple times in an invoice, due to significant annotation errors. 
This illustrates the difference between ``ground truth" when evaluating against relational-records rather than human annotations. Relational-records better reflect real-world objectives in most applications and this result suggests that human annotations are a rather poor proxy for evaluating these objectives.

Next, we examine the performance of DeepCPCFG which has been trained based on the compatible trees from the training data, as such it is penalized in this evaluation in the same way that the ``Leaf-annotations" are. Notably DeepCPCFG obtains results comparable to and sometimes better than the ``Leaf-annotations" on the test data.

From these experiments we see that the problem of information extraction is substantially different from the problem of classifying bounding boxes. Despite this we see that DeepCPCFG is quite effective at classifying bounding boxes even when trained without any hand annotations.

\begin{table}[tb]
    \centering
    \caption{F1 classification results on proprietary invoices evaluated on hand-annotations as ground truth}
    \label{tbl:hed-proprietary-rvl-cdip1}
    \begin{tabular}{|m{0.32\columnwidth}|L|Z|Z|Z|Z|J|Z|J|}
        \hline
        \multicolumn{1}{|c|}{Model} & \scriptsize{\bf Overall} & Desc & Qty & Rate & Amt & \scriptsize{InvoiceID} & Date & \scriptsize{TotalAmt} \\
        \hline 
        LayoutLM with token classification from Hand-annotations & 93.3 & 92.0 & 96.4 & 95.3 & 96.5 & 96.6 & 97.2 & 89.3 \\
        \hline
        Leaf-annotations & 86.4 & 95.6 & 87.9 & 61.3 & 65.4 & 88.8 & 72.2 & 52.7 \\
        \hline
        DeepCPCFG fine-tuned with LayoutLM using leaf-annotations (Epoch 1) & 81.4 & 88.1 & 87.3 & 58.2 & 64.0 & 88.7 & 75.9 & 52.1 \\
        \hline
    \end{tabular}
\end{table}

\section{Conclusion}
We have described and demonstrated a method for end-to-end structured information extraction from business documents such as invoices and receipts. This work enhances existing capabilities by removing the need for brittle post-processing, and by reducing the need for annotated document images.

Our method ``parses'' a document image using a grammar that describes the information to be extracted. The grammar does not describe the layout of the document or significantly constrain that layout. This method yields compelling results. However, research on this important problem is limited by the lack of available benchmark data sets which has slowed development and stymied comparisons. In order to alleviate this, we released a new public evaluation set based on the RVL-CDIP data. 
Related ``2D parsing'' approaches have been previously explored for image analysis and we believe that the effectiveness of our approach suggests a broader re-examination of grammars in image understanding particularly in combination with Deep Learning.


We have gone beyond existing work in this space by \cite{8892875,10.1109/ICDAR.2005.98,hwang2020spatial,subramani2020survey,layoutlm,yu2020pick}. The success of DeepCPCFG, together with earlier work in this space, shows the value of combining structured models with Deep Learning. 


Our experimental results illustrate the significant gap between information extraction and recovering labels from hand annotations. We believe that evaluations based on recovering relational-records best reflect real-world use cases. In ongoing work, we are applying this technique to a wide variety of structured business documents including tax forms, proofs of delivery, and purchase orders.

\subsubsection{Disclaimer} The views reflected in this article are the views of the authors and do not necessarily reflect the views of the global EY organization or its member firms.

\subsubsection{Acknowledgements} The authors will like to thank the following colleagues: David Helmbold, Ashok Sundaresan, Larry Kite, Chirag Soni, Mehrdad Gangeh, Tigran Ishkhanov and Hamid Motahari.

\bibliographystyle{splncs04}
\bibliography{pcfg}

\end{document}